# Graph Neural Network and Transformer Integration for Unsupervised System Anomaly Discovery

Yun Zi
Georgia Institute of Technology
Atlanta, USA

Ming Gong
University of Pennsylvania
Philadelphia, USA

Zhihao Xue
Rose-Hulman Institute of Technology
Terre Haute, USA

Yujun Zou
University of California, Berkeley
Berkeley, USA

Nia Qi
Independent Author
Pittsburgh, USA

Yingnan Deng *
Georgia Institute of Technology
Atlanta, USA

*Abstract*-This study proposes an unsupervised anomaly detection method for distributed backend service systems, addressing practical challenges such as complex structural dependencies, diverse behavioral evolution, and the absence of labeled data. The method constructs a dynamic graph based on service invocation relationships and applies graph convolution to extract high-order structural representations from multi-hop topologies. A Transformer is used to model the temporal behavior of each node, capturing long-term dependencies and local fluctuations. During the feature fusion stage, a learnable joint embedding mechanism integrates structural and behavioral representations into a unified anomaly vector. A nonlinear mapping is then applied to compute anomaly scores, enabling an end-to-end detection process without supervision. Experiments on real-world cloud monitoring data include sensitivity analyses across different graph depths, sequence lengths, and data perturbations. Results show that the proposed method outperforms existing models on several key metrics, demonstrating stronger expressiveness and stability in capturing anomaly propagation paths and modeling dynamic behavior sequences, with high potential for practical deployment.

## I. Introduction

In modern information infrastructure, distributed backend service systems serve as the core components of digital platforms. They are responsible for handling high-concurrency requests, task scheduling, and resource coordination. With the widespread adoption of cloud computing, microservice architectures, and container technologies, backend systems have become large-scale, structurally complex, and highly interactive. These systems are often composed of multiple heterogeneous nodes and services, which interact through dynamic dependencies to achieve collaborative operations. In this context, the operational state of the system relies heavily on the structural correlations and behavioral patterns among its components. Once a local node fails, it may trigger cascading failures, jeopardizing the stability and reliability of the entire service chain. Therefore, it is crucial to identify potential anomalies within the system accurately and to develop efficient, sensitive, and generalizable anomaly detection mechanisms. Such mechanisms are essential for ensuring the continuity and stability of distributed services.

Anomaly detection in distributed backend services faces multiple challenges. First, operational data such as system logs, monitoring metrics, and trace information are highly coupled in both temporal and structural dimensions. These data exhibit non-stationary behaviors, multi-source heterogeneity, and strong interdependencies. Traditional rule-based or single-sequence modeling approaches struggle to capture complex inter-service dependencies and global behavioral evolution, often resulting in false alarms or missed anomalies. Second, as system scale increases, the interaction patterns among service components evolve into graph-like structures. Anomalies manifest not only as sudden changes in individual metrics but also as propagations and local aggregations at the topological level. Such structural anomalies involve cross-node temporal dependencies and nonlinear dynamic evolutions, which are difficult to capture through static features or single-perspective modeling. Therefore, a composite modeling approach that jointly captures structural dependencies and behavioral sequences is critical for improving anomaly detection performance[1].

Graph Neural Networks (GNNs) have emerged as powerful deep learning tools for processing graph-structured data. They can model high-order relationships among multiple nodes and effectively capture information propagation patterns within topologies[2-4]. In distributed backend services, different service nodes and their interactions can form dynamic graphs. Through multi-layer aggregation mechanisms, GNNs can construct global structural representations based on local perceptions, enabling the identification of anomaly propagation paths and hidden inter-node impacts. Meanwhile, the Transformer architecture demonstrates strong capabilities in sequence modeling. Its attention mechanisms excel at capturing long-term dependencies and local fluctuations in time series data. Incorporating Transformers into system behavioral modeling enhances the expression of metric trends and facilitates the identification of key patterns in complex event evolutions. Integrating GNNs with Transformers enables the joint modeling of structural dependencies and temporal behaviors. This integration provides a more comprehensive and fine-grained representation foundation for anomaly detection in complex scenarios.

Furthermore, anomalies in distributed backend systems exhibit highly diverse forms. These include explicit anomalies such as metric drifts and latency spikes, as well as implicit anomalies such as subtle behavioral changes and structural perturbations. Most existing detection methods rely on predefined thresholds or supervisory signals. These approaches are insufficient in settings with limited or weak labeling. Given the scarcity of annotation resources and the extremely imbalanced distribution of anomaly categories, it is imperative to develop unsupervised models with adaptive learning capabilities[5]. By incorporating fused representations of structural and temporal information, models can reduce dependency on prior knowledge. This enhances generalization and robustness in detecting previously unseen anomalies, laying a theoretical and methodological foundation for building practical and deployable anomaly detection systems.

## II. RELATED WORK

Anomaly detection in distributed backend service systems has been extensively studied from both structural and temporal perspectives. Robust detection under real-world complexity often requires combining causal inference, dynamic graph modeling, and sequence learning. For instance, H. Wang proposes a causal discriminative framework for robust fault detection in cloud services, leveraging both correlation and causal paths to enhance the interpretability and resilience of anomaly discovery [6]. Building on multi-faceted behavioral signals, R. Meng et al. introduce federated contrastive learning for distributed anomaly detection, emphasizing the effectiveness of high-order representation alignment under privacy constraints [7].

Recent work also highlights the value of self-attention and feature fusion in unsupervised or weakly supervised anomaly scenarios. H. Xin and R. Pan explore self-attention-based modeling of multi-source metrics, demonstrating its ability to capture performance trends and subtle anomalies in cloud systems [8]. Similarly, Y. Ma presents a conditional multiscale GAN and adaptive temporal autoencoder approach for anomaly detection in microservice environments, which underscores the importance of multi-level sequence modeling and temporal reconstruction for capturing complex behavioral changes [9].

Beyond sequence learning, Z. Liu and Z. Zhang utilize deep Q-learning to model workflow dynamics, showing that reinforcement-driven structural modeling can help trace anomaly propagation paths and inform intelligent intervention strategies [10]. Directly addressing unsupervised scenarios, Y. Cheng proposes a selective noise injection and feature scoring method for request anomaly detection, which enables more robust representation learning and adaptive thresholding in the absence of labels [11].

Temporal modeling remains essential for proactive and generalizable detection. Y. Wang et al. combine time-series learning and deep neural architectures to achieve proactive fault prediction in distributed systems, effectively capturing temporal patterns associated with failure events [12]. Y. Ren furthers this direction by integrating structural encoding and multi-modal attention for root cause detection, demonstrating that hybrid models can boost both interpretability and detection sensitivity in complex topologies [13].

In cloud-native and resource-constrained settings, collaborative and multi-agent methods play a key supporting role. B. Fang and D. Gao propose a collaborative multi-agent reinforcement learning approach for elastic resource scaling, offering new perspectives on anomaly response and system robustness through intelligent control [14]. Knowledge fusion and multi-source information retrieval, as explored by Y. Sun et al., also facilitate more nuanced understanding of anomalous patterns by leveraging retrieval-augmented mechanisms for representation enrichment [15].

Finally, parameter-efficient model adaptation strategies are becoming increasingly important for practical deployment. X. Meng and colleagues design collaborative distillation strategies, enabling lightweight yet expressive language model deployment, which supports scalable and responsive anomaly detection across large distributed infrastructures [16]. Together, these methods—causal graph modeling, attention-based sequence learning, hybrid representation fusion, and scalable adaptation—form the methodological foundation for high-fidelity, unsupervised anomaly detection in dynamic distributed service systems.

## III. METHOD

This architecture diagram illustrates a joint modeling framework that integrates Graph Neural Networks and Transformers for anomaly detection in distributed backend services. The model first extracts structural dependency information among service nodes through multi-layer graph convolution. It then models the temporal behavior sequences of each node using a Transformer to capture long-term dependencies and local fluctuations. The structural and behavioral representations are fused through an embedding integration module, and a nonlinear mapping is applied to generate anomaly scores. This enables efficient detection of abnormal states in complex systems. The model architecture is shown in Figure 1.

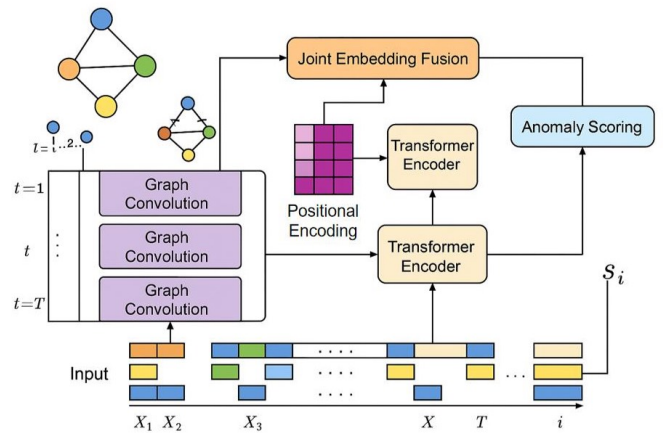

Figure 1. GTF-Net overall model architecture

The method in this paper aims to build a joint modeling framework that integrates graph structure information and temporal behavior features to improve the accuracy and generalization of distributed backend service anomaly detection. First, the backend service system is modeled as a dynamic graph structure, where each service node is represented as a vertex in the graph, and the calls or dependencies between services are represented as edges. Given a system snapshot at a time step $t$, the corresponding graph representation $G^{(t)} = (V^{(t)}, E^{(t)})$ can be constructed, where $V^{(t)}$ represents the node set and $E^{(t)}$ represents the edge set. Each node $v_i \in V^{(t)}$ corresponds to a set of time series features $Xi = [x_i^1, x_i^2, ..., x_i^T]$, where $x_i^t \in R^d$ represents the observation value at time $t$. To capture the structural information between nodes, the graph convolution operation is used to update the node representation, which is specifically expressed as:

$$H^{(l+1)} = \sigma(\tilde{D}^{-1/2} \tilde{A} \tilde{D}^{-1/2} H^{(l)} W^{(l)})$$

$\tilde{A} = A + I$ is the adjacency matrix after adding self-loops, $\tilde{D}$ is the corresponding degree matrix, $W^{(l)}$ is the weight parameter of the $l$ th layer, $H^{(l)}$ is the node representation of the $l$ th layer, and $\sigma$ is the activation function. The graph neural network part is used to extract the potential dependencies in the topological structure between services and form a high-dimensional embedding representation at the structural level.

To further model the evolution of the behavior sequence of each node, this paper introduces a Transformer-based temporal encoding module [17-18]. The time series features of each node are taken as input, and the temporal order information is introduced through positional encoding. Suppose the input sequence of a node is $X_i \in R^{T \times d}$, and its corresponding sequence embedding is expressed as:

$$Z_i = TransformerEncoder(X_i + P)$$

Where $P \in R^{T \times d}$ represents the position encoding matrix, and $Z_i \in R^{T \times d}$ is the embedding result after temporal encoding. Through the multi-layer self-attention mechanism, the model can capture long-term dependency features and local mutation patterns, and improve the ability to represent abnormal behaviors. The calculation formula for attention weight is as follows:

$$Attention(Q, K, V) = soft\max(\frac{QK^T}{\sqrt{d_k}})V$$

Where $Q, K, V$ are the query, key, and value matrices, respectively, obtained by linear mapping of the input sequence, and $d_k$ is the scaling factor of the attention dimension. This mechanism enables the model to dynamically allocate attention between different time steps, thereby identifying key variation patterns in the behavior sequence.

To achieve the effective fusion of structural information and behavioral sequence features, this paper designs a joint embedding fusion mechanism in the output stage. Assuming that the output of the graph neural network is $H_i \in R^d$ and the Transformer encoding result is $Z_i^{final} \in R^d$, the final node comprehensive representation is:

$$U_i = \gamma H_i + (1 - \gamma) Z_i^{final}$$

$\gamma \in [0,1]$ is a learnable fusion weight parameter, which represents the importance ratio of structure and time series information. Finally, the comprehensive representation $U_i$ is transformed through a nonlinear mapping function to obtain the abnormality score of each node:

$$s_i = f(U_i) = MLP(U_i)$$

$MLP$ represents a multi-layer perceptron structure, which is used to complete the nonlinear mapping from the embedding space to the abnormal space. This score will be used to measure the abnormality of the node at the current time, providing refined state assessment support for the system. The overall method is trained under unsupervised conditions and identifies abnormal behaviors based on the degree of deviation in the representation space.

IV. EXPERIMENTAL RESULTS

*A. Dataset*

This study uses the Alibaba Cluster Trace Program (Ali-CEP) dataset, publicly released by the Alibaba Cloud AIOps platform, as the evaluation data for anomaly detection in distributed backend services. The dataset includes operational metrics, task scheduling records, and service dependency information from multiple nodes in a large-scale cluster. It reflects the dynamic evolution of backend systems in a real cloud computing environment. The dataset contains historical behavior sequences from over 1000 service instances, covering multi-dimensional metrics such as CPU usage, memory consumption, disk I/O, and network traffic. It also provides topological dependency information among services, showing clear graph-structured characteristics.

The monitoring metrics are sampled at minute-level intervals. This provides high temporal resolution and long observation periods, which support the modeling of long-term behavioral evolution and short-term fluctuations. By extracting service invocation relationships, dynamic multi-hop service graphs can be constructed to capture structural anomaly propagation patterns. The dataset also includes a small number of system alerts and scheduling anomalies, which can be used as weak labels for auxiliary evaluation. However, the core detection task remains unsupervised.

The high-dimensional, multi-source, and graph-structured nature of the Ali-CEP dataset aligns well with the modeling approach proposed in this study. It supports the joint modeling of structural dependencies and temporal behaviors. The dataset offers a solid foundation for evaluating the method's effectiveness. It is particularly suitable for simulating diverse anomaly patterns in complex distributed backend service environments.

### B. Experimental Results

This paper first conducts a comparative experiment, and the experimental results are shown in Table 1.

Table1. Comparative experimental results

| Model | F1 Score | AUC | KS Score | Time Cost |
|---|---|---|---|---|
| **Ours (GTF-Net)** | 0.889 | 0.942 | 0.741 | 19.6 ms |
| **GDN[19]** | 0.841 | 0.913 | 0.695 | 17.4 ms |
| **Anomaly Transformer [20]** | 0.864 | 0.927 | 0.714 | 21.1 ms |
| **DONUT+ [21]** | 0.798 | 0.884 | 0.621 | 15.7 ms |
| **MO-GAAL [22]** | 0.781 | 0.857 | 0.59 | 36.2 ms |

Experimental results demonstrate that the proposed GTF-Net outperforms existing state-of-the-art models across all evaluation metrics, confirming its effectiveness in anomaly detection for distributed backend services. Specifically, GTF-Net achieves an F1 Score of 0.889 and an AUC of 0.942, both significantly higher than those of the baseline models. These results indicate that the model offers stronger overall performance in balancing precision and recall. In particular, GTF-Net reaches a KS Score of 0.741, suggesting superior discriminative power in separating normal and anomalous distributions. This indicates its robustness in handling various anomaly patterns.

Among the baseline models, Anomaly Transformer shows strong performance in temporal modeling. Its self-attention mechanism is effective in capturing long-range dependencies. However, it does not explicitly model the structural relationships among services, which leads to lower detection accuracy in scenarios with strong structural dependencies. GDN, as a graph-based anomaly detection model, performs well in capturing static topological features among services. Nevertheless, it cannot model dynamic behavioral sequences, making it less effective in detecting anomalies that evolve. As a result, its performance is slightly inferior to that of the proposed method.

DONUT+ and MO-GAAL represent unsupervised paradigms based on reconstruction and adversarial learning, respectively. However, both are primarily designed for univariate or weakly structured data scenarios. They struggle to adapt to multi-source, high-dimensional, and tightly coupled backend service environments. In particular, MO-GAAL shows limited generalization in complex distributed systems with highly interactive nodes. Its KS and F1 scores are lower than those of other models, confirming the performance limitations caused by the lack of structural modeling constraints.

Overall, GTF-Net introduces a graph neural network to capture the topological structure among services, while integrating a Transformer to dynamically encode the temporal behaviors of nodes. This joint modeling approach enables the detection of complex anomalies such as localized propagation and cross-node behavior drift. The hybrid design enhances the model's capacity to recognize collaborative anomalies in distributed environments. Experimental results validate the model's comprehensive advantages in accuracy, robustness, and discriminative power, highlighting its practical potential for real-world system operation and maintenance.

This paper also experiments on the sensitivity of Transformer encoding depth to the temporal modeling effect. The experimental results are shown in Figure 2.

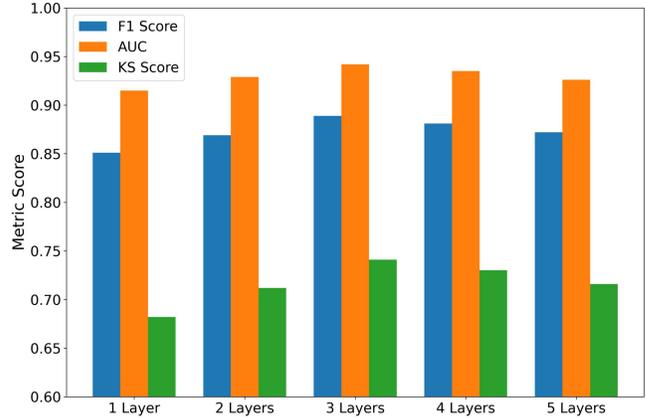

Figure 2. Sensitivity analysis of Transformer encoding depth on temporal modeling effect

Experimental results show that the number of Transformer encoder layers has a significant impact on the model's anomaly detection performance. When the number of layers is small, such as one or two, the model has limited capacity to capture complex dependencies in time series. This results in relatively low scores across various metrics, especially in F1 Score and KS Score, which fail to reflect the full characteristics of anomalous behaviors. These findings suggest that shallow architectures are insufficient for capturing long-term behavior evolution and subtle anomaly signals.

When the number of Transformer layers increases to three, the model achieves optimal performance. F1 Score, AUC, and KS Score all show marked improvements. At this depth, the model reaches a good balance between structural complexity and feature representation. The multi-layer self-attention mechanism enhances the ability to capture contextual patterns before and after anomaly occurrences. This improves the model's capacity to identify latent deviations and perturbations in service behaviors. Further increasing the number of layers to four and five leads to a slight performance decline. Although overall performance remains high, deeper stacking introduces risks of redundant modeling and overfitting. This is especially evident in high-dimensional and sparse time series, where attention weight distributions become diluted and key signals are harder to focus on. Excessive depth also results in training instability and higher computational costs, which can affect deployment efficiency. This experiment confirms the sensitivity of Transformer depth in modeling temporal behaviors in distributed backend services. It highlights the importance of choosing an appropriate depth for capturing service state evolution and temporal anomalies. Conclusion

This study addresses the challenges of structural complexity, behavioral diversity, and label scarcity in anomaly detection for distributed backend services. It proposes an unsupervised modeling method that integrates Graph Neural Networks and Transformer architectures. The method leverages topological information among system components and the temporal evolution of node behaviors. Through the synergy of graph-based modeling and attention mechanisms, it enables efficient detection of latent anomaly patterns in multi-dimensional monitoring data. The model integrates structural dependencies and temporal features at the representation level. This enhances sensitivity to cross-node anomaly propagation and local behavioral shifts, offering a robust solution for anomaly detection in complex environments.

Applied to real-world monitoring data from cloud computing platforms, the method demonstrates strong performance in both accuracy and stability. It shows superior generalization and representational power, especially in service systems characterized by strong coupling and high volatility. Compared with traditional approaches that rely solely on local metric modeling, the proposed method adopts a joint structural and sequential perspective. This improves the model's capacity to recognize heterogeneous anomaly patterns and reduces dependency on predefined rules and large-scale labels. It provides a solid foundation for intelligent awareness and proactive alerting in operational systems. The proposed framework applies not only to backend service anomaly detection in cloud and big data platforms but also to other graph-structured temporal environments. These include industrial Internet of Things, financial risk control, and distributed edge computing scenarios. Its general modeling paradigm of combining structure and behavior supports unsupervised learning in large-scale heterogeneous systems. It offers both theoretical and practical guidance for enhancing observability, interpretability, and autonomy in complex systems.

## V. Future work

Future work may explore the framework's adaptability to multimodal systems. By incorporating heterogeneous information sources such as log texts, event sequences, and network topologies, it is possible to construct a multi-perspective anomaly semantic space. Under large-scale deployment, challenges such as incremental online learning, cross-domain adaptation, and federated distributed implementation will become critical. Introducing multi-task self-supervision and cross-modal representation alignment can lead to more intelligent and adaptive anomaly detection systems. This will support the evolution of distributed services toward autonomous and intelligent operations.